\documentclass[journal]{IEEEtran}
\usepackage{graphicx}
\graphicspath{{figures/}}
\usepackage{url}
\usepackage{array}
\usepackage{amsmath}
\usepackage{amssymb}
\usepackage{multirow}
\usepackage{multicol}
\usepackage{booktabs}
\usepackage{colortbl}
\usepackage{cite}
\usepackage{algorithm}  
\usepackage{algorithmicx}  
\usepackage{algpseudocode}  
\usepackage{amsmath}  
\usepackage{amsfonts,amssymb}  
\UseRawInputEncoding
\floatname{algorithm}{Algorithm}

\usepackage[colorlinks=true, bookmarks=false, citecolor=black,
           linkcolor=black,
           anchorcolor=black,
           urlcolor=black 
           ]{hyperref}

\usepackage{xspace}
\makeatletter
\DeclareRobustCommand\onedot{\futurelet\@let@token\@onedot}
\def\@onedot{\ifx\@let@token.\else.\null\fi\xspace}

  \def\ie{\emph{i.e}\onedot}
  
\def\etc{\emph{etc}\onedot}  
 \def\etal{\emph{et al}\onedot}
\makeatother

\newcommand{\PreserveBackslash}[1]{\let\temp=\\#1\let\\=\temp}
\newcolumntype{C}[1]{>{\PreserveBackslash\centering}p{#1}}
\newcolumntype{R}[1]{>{\PreserveBackslash\raggedleft}p{#1}}
\newcolumntype{L}[1]{>{\PreserveBackslash\raggedright}p{#1}}

\newcommand{\tabincell}[2]{\renewcommand{\arraystretch}{1.3}\begin{tabular}{@{}#1@{}}#2\end{tabular}}

\DeclareMathSizes{10}{10}{6}{5}

\begin{document}
\title{ReLoc: A Restoration-Assisted Framework for Robust
Image Tampering Localization}
\author{
    Peiyu~Zhuang,
    Haodong~Li,
	Rui~Yang,
    Jiwu~Huang,
    \thanks{Corresponding author: Haodong Li.}
   \thanks{P. Zhuang, H. Li, and J. Huang are with the Guangdong Key Laboratory
   of Intelligent Information Processing and Shenzhen Key Laboratory of Media
   Security, Shenzhen University, Shenzhen 518060, China; and also with the
   Shenzhen Institute of Artificial Intelligence and Robotics for Society,
   Shenzhen 518060, China. (email: 1800261051@email.szu.edu.cn; lihaodong,
   jwhuang@szu.edu.cn)}
   \thanks{R. Yang is with the Alibaba Group, Hanzhou 311121, China. (email:
   duming.yr@alibaba-inc.com) }
}
\maketitle

\begin{abstract}
With the spread of tampered images, locating the tampered regions in digital
images has drawn increasing attention. The existing image tampering localization
methods, however, suffer from severe performance degradation when the tampered
images are subjected to some post-processing, as the tampering traces would be
distorted by the post-processing operations. The poor robustness against
post-processing has become a bottleneck for the practical applications of image
tampering localization techniques. 
In order to address this issue, this paper proposes a novel
\textit{re}storation-assisted framework for image tampering
\textit{loc}alization (ReLoc). The ReLoc framework mainly consists of an image
restoration module and a tampering localization module. The key idea of ReLoc is
to use  the restoration module to recover a high-quality counterpart of the
distorted tampered image, such that the distorted tampering traces can be
re-enhanced, facilitating the tampering localization module to identify the
tampered regions. 
To achieve this, the restoration module is optimized not only with the
conventional constraints on image visual quality, but also with a
forensics-oriented objective function. Furthermore, the restoration module and
the localization module are trained alternately, which can stabilize the
training process and is beneficial for improving the performance. The proposed
framework is evaluated by fighting against JPEG compression, the most commonly
used post-processing. Extensive experimental results show that ReLoc can
significantly improve the robustness against JPEG compression. The restoration
module in a well-trained ReLoc model is transferable. Namely, it is still
effective when being directly deployed with another tampering localization
module.
\end{abstract}


\IEEEpeerreviewmaketitle
\section{Introduction}
\label{Sec:introduction}
\begin{figure}[htbp]
    \centering
    \includegraphics[width=\columnwidth]{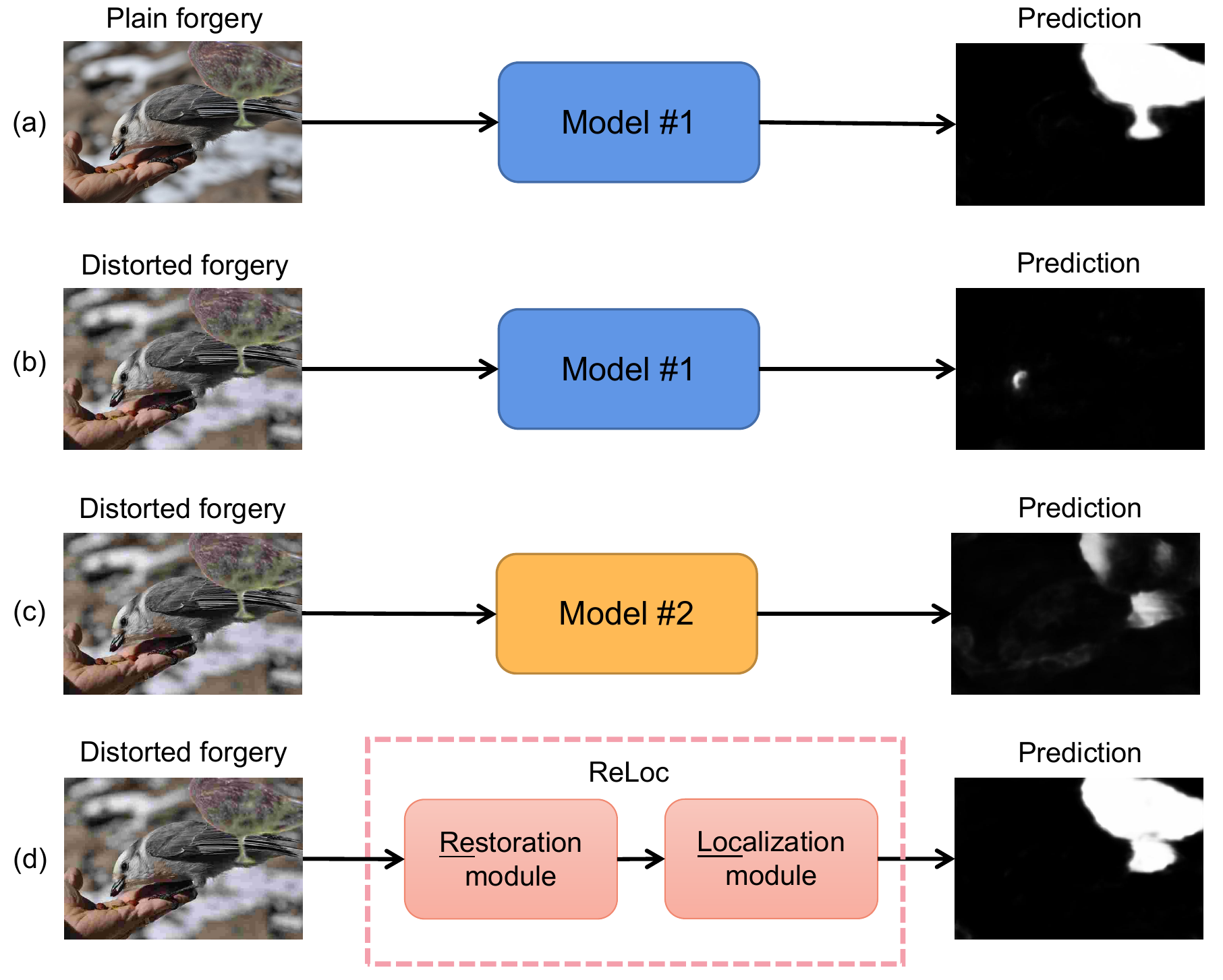}
    \caption{Tampering localization results in different situations.
    The localization model \#1 was trained with plain tampered images, model \#2 was
    fine-tuned with distorted images based on model \#1, and the localization module
    in ReLoc was fine-tuned with restored images.}
    \label{fig:before_after_compression}
\end{figure}

\IEEEPARstart{A}{s} an important carrier of information transmission, digital
images appear widely in our daily life. With the development of image processing
technology, people can easily create realistic tampered images by using various
image editing software. Once the tampered images are used for malicious purposes,
such as forging certificates, creating rumors, \etc, it is bound to result in a
series of negative impacts. Therefore, in order to prevent the abuse of tampered
images, it is of great significance to identify tampered images.

In recent years, many forensic methods have been proposed to detect tampered
images and further localize the tampered
regions \cite{verdoliva2020media,castillo2021comprehensive}. The technological
paradigms of these methods have shown a trend from relying on hand-crafted
features  \cite{Mayer2018AccurateAE,Gallagher2008ImageAB,Yuan2011BlindFO} to
utilizing deep learning techniques \cite{salloum2018image,
sun_et_2022,zhong2020endtoend, cozzolino2015efficient,wu2022robust,
li2019localization, wu2022iidnet,hu2020span, wu2019mantranet, liu2021psccnet,
zhuang2021image,hao_transforensics_2021,wang_objectformer_2022}. Nowadays, the
deep learning (DL)-based forensic methods usually achieve much better
performance than the conventional ones. For one thing, this can be attributed to the use
of some effective network architectures, such as fully convolutional
network \cite{long2015fully}, U-Net \cite{ronneberger2015unet}, faster
R-CNN \cite{ren2015faster}, mask-RCNN \cite{he2017mask} and ViT
 \cite{dosovitskiy_image_2021}. For another thing, the combination of deep
learning and domain knowledge in forensics also plays an important role
 \cite{fridrich2012rich, bayar2018constrained}.

Although the existing tampering localization methods can achieve good
performance on some datasets in laboratorial evaluations, in practical
applications they suffer from severe performance degradation when the tampered
images are subjected to a series of post-processing operations, such as JPEG
compression, blurring, scaling, \etc. Because post-processing would seriously
distort the tampering traces. For conciseness, hereinafter we refer the tampered
images without post-processing as \emph{plain images} and the post-processed
counterparts as \emph{distorted images}. The phenomenon mentioned above can be
intuitively interpreted with Fig.~\ref{fig:before_after_compression}. Generally,
a tampering localization model is trained with a set of plain images. The model
works well if the investigated image is a plain forgery
(Fig.~\ref{fig:before_after_compression}-a). However, if a distorted tampered
image is fed to the model, the tampered regions are likely failed to be
identified (Fig.~\ref{fig:before_after_compression}-b). As a result,
post-processing has become a stumbling stone for the practical applications of
image tampering localization.

In order to improve the robustness against post-processing, some recent works
have tried to introduce distorted images in the training phase. Rao \etal
\cite{rao2021selfsupervised} proposed to learn robust features from JPEG proxy
images. They employed a JPEG proxy network to simulate the process of JPEG
compression and produce JPEG proxy images. The proxy images were then used to
train the localization model. To obtain a robust model against online social
networks (OSN), Wu \etal \cite{wu2022robust} firstly simulated the noise
introduced by OSN and then introduce the noise into the training images to
generate distorted images. However, since the tampering traces have been
distorted by post-processing, it is difficult to learn discriminative features.
As shown in Fig.~\ref{fig:before_after_compression}-c, even though the
localization network was fine-tuned directly with distorted images, the tampered
regions cannot be well detected.

Different from the existing approaches, this paper tries to improve the
robustness against post-processing from a novel perspective. Our key idea is
that, once the tampering traces distorted by post-processing can be recovered or
re-enhanced, it is able to learn effective representations for tampering
localization. To this end, we propose a \emph{re}storation-assisted framework
for robust image tampering \emph{loc}alization (ReLoc). ReLoc is a hybrid
framework that cascades a restoration module and a localization module
(Fig.~\ref{fig:before_after_compression}-d). The restoration module aims to
re-enhancing the distorted tampering traces and recovering a high-quality
counterpart from the distorted image. The localization module takes the restored
images as input, so that the subtle tampering traces would be captured more
efficiently. We utilize a pixel-level loss, an image-level loss, and a
forensics-oriented localization loss to train the restoration module. In this
way, the restoration module is jointly constrained by image visual quality and
tampering localization efficacy. We also propose to optimize the restoration and
localization modules in an alternate way, which can make the training process
more stable and achieve better performance. To evaluate the effectiveness of
ReLoc, we consider a typical case on handling against JPEG compression, which is
one of the most commonly used post-processing operations. The extensive
experiments involving three tampering localization methods on three different
datasets show that ReLoc can significantly improve the robustness against JPEG
compression, regardless of whether the compression quality is fixed or
changeable. Another merit of ReLoc is that the restoration module in a
well-trained model is transferable, meaning that it can be directly deployed
with another localization module for improving the robustness of tampering
localization.

The main contributions of this work are as follows.
\begin{itemize}
	\item We propose a new idea to improve the robustness of tampering
    localization against post-processing. That is, to restore the distorted
    image before performing tampering localization. Based on this idea, we
    propose a restoration-assisted tampering localization framework (ReLoc). By
    utilizing the restoration module, the distorted tampering traces can be
    recovered to some extent, so that the tampering localization module can
    capture tampering traces more efficiently. Consequently, the robustness will
    be improved. To our best knowledge, this is the first work to utilize image
    restoration to achieve robust tampering localization.

    \item We design a training schema for ReLoc that is tailored for image
    tampering localization. To train the restoration module, in addition to
    considering image visual quality, we also include a forensics-oriented loss
    for ensuring tampering localization performance. To make the training
    process stable, we optimize the restoration and localization modules in an
    alternate way. The experimental results show that the proposed method can
    effectively improve the robustness against post-processing.

    \item We provide a plug-and-play restoration module for robust tampering
    localization via introducing the ReLoc framework. We experimentally validate
    that the restoration module in a well-trained ReLoc model is also effective
    to work with another localization module, meaning that ReLoc can be flexibly
    deployed in practical applications. 
\end{itemize}

The rest of this paper is organized as follows. Section~\ref{Sec:related_work}
reviews the related works on image restoration and image tampering localization.
Section~\ref{Sec:proposed_method} describes the proposed framework in detail.
Section~\ref{Sec:experiments} presents the experimental results and discussions.
Finally, Section~\ref{Sec:conclusion} draws the concluding remarks.

\section{Related Work}
\label{Sec:related_work}
\subsection{Image Restoration} 

As a fundamental problem in image processing, image restoration aims to recover
visually pleasant high-quality images from degraded low-quality images. Usually,
the degraded images are subjected to different distorted operations, such as
down-sampling, blurring, and lossy compression. As a result, image restoration
includes denoising, super-resolution, deblurring, \etc. Image restoration is an
ill-posed inverse problem. To solve such a problem, conventional methods usually
resort to certain image priors and mathematical models
\cite{chantas2010variational,chantas2006bayesian,molina2001image}. Recently, the
research of image restoration has been dominated by DL-based methods, as deep
learning is very effective for improving image visual quality.

The DL-based image restoration methods usually employ convolutional neural
network (CNN) \cite{krizhevsky2012imagenet} or Transformer
\cite{vaswani_attention_2017} as basic architectures. Dong \etal
\cite{dong2014learning} proposed the first CNN-based image super-resolution
method, in which the model was constructed by stacking convolutional layers.
They also applied CNN to the reduction of JPEG compression artifacts
\cite{dong2015compression}. By adopting residual learning, Kim \etal
\cite{kim2016accurate} proposed a method for image super-resolution. In order to
make use of the hierarchical features extracted at different levels in CNNs,
Zhang \etal \cite{zhang2021residuala} proposed a residual dense network for
image super-resolution, denoising, and deblurring. Recently, with the promising
performance achieved by Vision Transformer \cite{dosovitskiy_image_2021},
Transformer-based methods have become popular in the field of image restoration,
such as U-Former \cite{wang2021uformera} and SwinConv-Unet
\cite{zhang2022practical}. 

In addition to network architectures, the objective
functions also play an important role in image restoration. While pixel-level
MAE and MSE losses are commonly used for optimization, they tend to produce
smooth images. To alleviate this issue, the perceptual loss
\cite{johnson2016perceptual} is used to make the restoration images more
pleasant for human eyes.

Since image restoration can recover the information loss caused by degradation
to a certain extent, it is beneficial for some high-level vision applications,
such as image classification, semantic segmentation, and object detection, in
the case that the given images are distorted ones. Haris \etal
\cite{haris2021taskdriven} proposed to optimize a super-resolution network for
object detection in low-resolution images. To improve the performance of
semantic segmentation for distorted images, Niu \etal \cite{niu2020effective}
applied denoising and super-resolution to the input images through a restoration
network. In this paper, we build a bridge between image restoration and image
forensics. To our best knowledge, this is the first work to show that
incorporating an image restoration module with a tampering localization module
can significantly improve the tampering localization performance for distorted images.

\begin{figure*}[htbp]
	\centering
	\includegraphics[width=\textwidth]{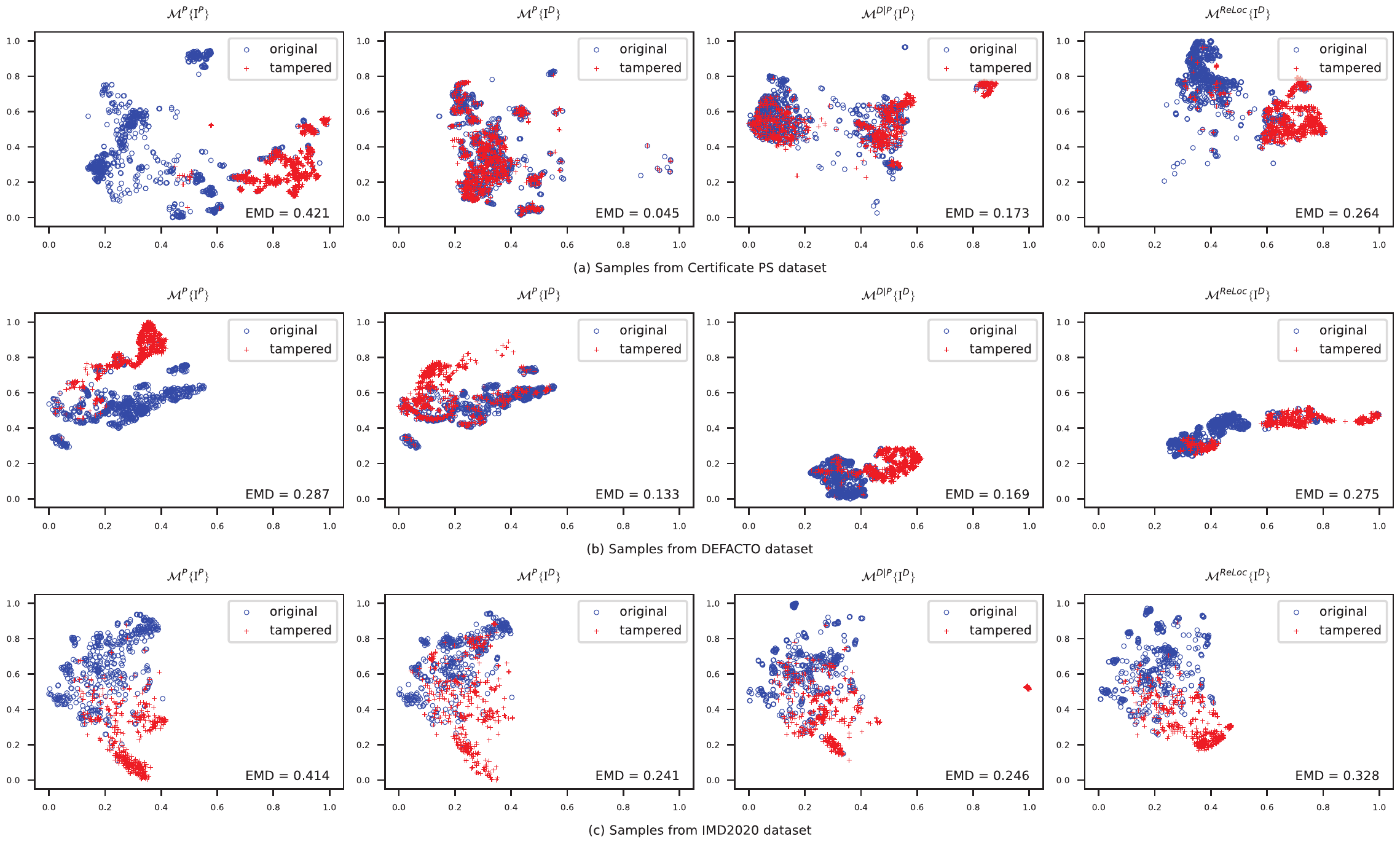}
	\caption{The feature representations of 1,000 original pixels and 1,000
		tampered pixels projected by T-SNE in different situations. The earth
		mover's distance (EMD) is used to measure the distance between the
		distributions of the two types of pixels.}
	\label{fig:Distribution}
\end{figure*}

\subsection{Image Tampering Localization}
Image tampering detection and localization methods are in urgent need as the
increasing abuse of tampered images in our daily life. With the development of
deep learning, many DL-based tampering detection and localization methods have
been proposed \cite{castillo2021comprehensive}. Some of them are targeted for a
certain tampering operation, such as splicing \cite{salloum2018image,
sun_et_2022}, copy-move \cite{wu2018busternet}, and inpainting
\cite{li2019localization,wu2022iidnet}. On the other hand, some are designed for
general detection. Wu \etal \cite{wu2019mantranet} trained a feature extractor
with 385 types of tampering operations, and then built a local anomaly detection
network based on LSTM to predict the tampered regions. Liu \etal
\cite{liu2021psccnet} proposed a PSCC-Net, which made use of the attention
mechanism \cite{wang2018non} and performed tampering localization at different
scales through a progressive network structure. Chen \etal \cite{chen2021image}
proposed a MVSS-net for tampering localization by leveraging the multi-view and
multi-scale information. To capture the subtle tampering traces, Zhuang \etal
\cite{zhuang2021image} designed a network with the dense block
\cite{huang2017densely} and included the commonly used operations in Photoshop
to build a dataset for pre-training. Kwon \etal \cite{kwon2021catnet} performed
tampering localization with dual-domain information. They used a DCT branch to
capture JPEG compression traces and fused the features extracted from a spatial
branch. As Transformer \cite{vaswani_attention_2017} can better model the global
information, some tampering localization methods based on Transformer have also
been developed, such as TransForensics \cite{hao_transforensics_2021} and
ObjectFormer \cite{wang_objectformer_2022}. Despite the above methods can obtain
good performance on some datasets, they did not address the robustness against
post-processing through an explicitly algorithmic design. As a result, their
performance would be significantly degraded in practical situations.

To improve the robustness against post-processing, some efforts have been
conducted. For example, Abecidan \etal \cite{abecidan2021unsupervised} improved
the robustness to JPEG compression by using domain adaptation between the source
and target domains, so that the detector was constrained to learn a robust
feature representation. Rao \etal \cite{rao2021selfsupervised} simulated the
impact of JPEG compression by generating proxy images, which can facilitate the
learning of tampering traces when the images are undergone JPEG compression. Wu
\etal \cite{wu2022robust} considered handling the tampering images that are
transmitted through online social networks. They firstly estimated the noise
introduced by OSN transmission. Then, they augmented the tampered images with
the predicted noise and trained the detector to make it robust to online
transmission. Although these works explicitly consider post-processing
operations, the detectors are difficult to learn effective features from the
distorted images, due to the fact that the tampering traces have been weakened by
post-processing. Different from them, we design a new framework to address the
problem of robustness. In the proposed method, a given distorted image is
processed by a restoration module before being fed to a tampering localization
module. In this way, the tampering traces can be recovered to some extent,
promoting the localization module to learn more robust features.


\section{Proposed method}
\label{Sec:proposed_method}

In this section, we elaborate on the proposed restoration-assisted framework for
robust image tampering localization (ReLoc). We first analyze the feature
representation under different situations, so as to give a better understanding
of the intrinsic mechanism of the robustness problem. After that we give an
overview of ReLoc, and then describe how to design a forensics-oriented
restoration task and how to optimize the overall framework.

To make the descriptions concise in the following contexts, we denote a plain
image, a distorted image, and a restored image as $\mathbf{I}^P$,
$\mathbf{I}^D$, and $\mathbf{I}^R$, respectively. We also denote a model as
$\mathcal{M}$ and use a superscript to indicate its training situation.
Namely, $\mathcal{M}^{P}$ and $\mathcal{M}^{D|P}$ mean a model trained with a set of 
$\mathbf{I}^P$ and a model further fine-tuned with a set of $\mathbf{I}^D$, respectively.
For a model built with ReLoc, it is denoted as $\mathcal{M}^{ReLoc}$, and the
restoration module and localization module are denoted as
$\mathcal{M}^{ReLoc}_R$ and $\mathcal{M}^{ReLoc}_L$, respectively. To describe a
testing situation that a set of image $\mathbf{I}^*$ is fed to a trained model
$\mathcal{M}^*$, we refer to $\mathcal{M}^{*}\{\mathbf{I}^*\}$. In this way, the
four situations shown in Fig.~\ref{fig:before_after_compression} are denoted as
$\mathcal{M}^{P}\{\mathbf{I}^P\}$, $\mathcal{M}^{P}\{\mathbf{I}^D\}$,
$\mathcal{M}^{D|P}\{\mathbf{I}^D\}$, and $\mathcal{M}^{ReLoc}\{\mathbf{I}^D\}$,
respectively.

\subsection{Analysis of Feature Representations}

In order to improve the robustness against post-processing, a straightforward
approach is to introduce the corresponding post-processing operations in the
training phase. Namely, to train a model with distorted images and expect that
it can learn robust feature representation. However, as the tampering traces in
the distorted images have been weakened by post-processing, it is tricky for the
tampering localization model to learn effective features directly from distorted
images.

To better interpret what is mentioned above, we analyze the feature
representations between tampered pixels and original pixels in different
situations. Firstly, we trained tampering localization models based on the
network architecture proposed in \cite{wu2022robust}\footnote{Similar results
can be obtained by using other tampering localization methods, such as
\cite{zhuang2021image,chen2021image}.}. By using the trained models, we then
tested the tampered images in the Certificate PS dataset (self-created, refer to
Section~\ref{subsubSec:dataset} for details), DEFACTO dataset
\cite{mahfoudi2019defacto}, and IMD2020 dataset \cite{novozamsky2020imd2020},
respectively. We randomly selected 1000 original and 1000 tampered pixels from
the tampered images in each dataset and used T-SNE \cite{van2008visualizing} to
project their feature representations extracted by the encoder of the trained
models to a 2-D space. The scatter plots of the 2-D representations are shown in
Fig.~\ref{fig:Distribution}.   
As shown in the leftmost column of Fig.~\ref{fig:Distribution}, in the case 
$\mathcal{M}^{P}\{\mathbf{I}^P\}$ (\ie, a model trained with plain images is
used to test plain images), the distribution gaps between the representations of
tampered pixels and original pixels are large. At this time, the model can well
localize the tampered regions. However, when the tampered images are subjected
to lossy post-processing, the tampering traces would be distorted. As shown in
the second column of Fig.~\ref{fig:Distribution}, the distributions of original
pixels and tampered pixels become confused, and the distances between the
distributions of the two types of pixels are significantly reduced. In this
case, the performance of the localization model will be very poor, as shown in
Fig.~\ref{fig:before_after_compression}-b.

In order to improve the tampering localization performance for distorted images,
we can fine-tune the model $\mathcal{M}^{P}$ with distorted images and obtain a
fine-tuned model $\mathcal{M}^{D|P}$, and then test $\mathbf{I}^D$ with
$\mathcal{M}^{D|P}$. In this case, as shown in the third column of
Fig.~\ref{fig:Distribution}, the distances between the distributions of original
pixels and tampered pixels increase compared to the case 
$\mathcal{M}^{P}\{\mathbf{I}^D\}$. However, it is observed that the two
distributions are still not as distinguishable as that in the case 
$\mathcal{M}^{P}\{\mathbf{I}^P\}$. Based on these observations, we conclude
that it is difficult to improve the robustness of tampering localization
methods by simply training on distorted images. A possible reason is that the
post-processing operations have inevitably led to information loss on the
tampering traces, and the localization model is difficult to learn effective
feature representations for the tampering traces from distorted images due to
information loss. To alleviate this issue, there is a need to recover the
information loss caused by post-processing.

\begin{figure*}[htbp]
	\centering
	\includegraphics[width=0.95\textwidth]{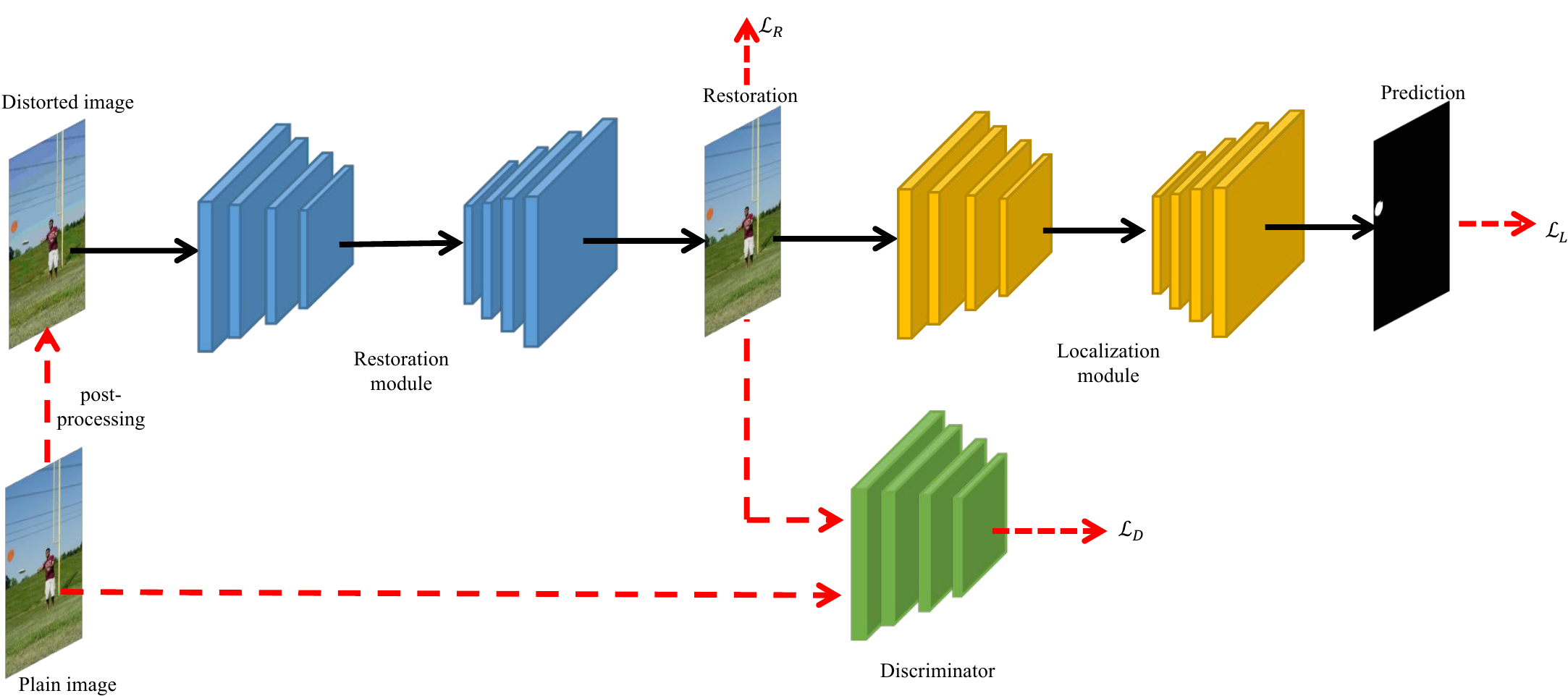}
	\caption{The diagram of the proposed ReLoc framework. Note that the process
	represented by the red dotted lines are only applicable for the training
	process. The restoration and localization modules are shown with
	encoder-decoder structures in this figure, but they are not limited to such
	a specific type of structure.}
	\label{fig:BaseModel}
\end{figure*}

\subsection{Overview of ReLoc}

Based on the above analysis, we are aware of that the information loss
introduced by post-processing increases the difficulty for distinguishing the
tampered pixels from the original pixels. Consequently, we propose a new idea to
improve the robustness against post-processing. Namely, restoring the distorted
images first and then performing tampering localization on the restored images.
It is expected that the information loss would be remedied via restoration, so
that the tampering localization performance can be further improved. Based on
this idea, we propose a restoration-assisted image tampering localization
framework, named ReLoc. As shown in Fig.~\ref{fig:BaseModel}, ReLoc mainly
consists of a restoration module, a localization module, and a discriminator.
For a given distorted image, the restoration module takes it as input and
produces a restored image. Then, the tampering localization module infers the
tampered regions by using the restored image. The discriminator, aiming to help
the restoration module generate better restored images, is utilized only in the
training phase. To make the restoration effective, we design different loss
functions to optimize the restoration network, including the pixel-level and
image-level losses, as well as a loss regarding to forensic performance
(Section~\ref{subsec:loss}). To make the whole framework be trained in a stable
way, we develop an alternate training strategy for the restoration module and
the localization module (Section~\ref{subsec:opt}). As shown in the rightmost
column of Fig.~\ref{fig:Distribution}, by using ReLoc, the distances between the
distributions of original pixels and tampered pixels become larger than the
cases of $\mathcal{M}^{D|P}\{\mathbf{I}^D\}$ and
$\mathcal{M}^{P}\{\mathbf{I}^D\}$, implying that ReLoc can effectively improve
the robustness of tampering localization against post-processing.

\subsection{Forensics-oriented Restoration}
\label{subsec:loss}


As we know, the degradation process is irreversible and image restoration is an
ill-posed inverse problem, meaning that it is unable to completely transform a
distorted image to its plain version via restoration. To restore a low-quality
image to a visually pleasant high-quality image, existing image restoration
methods utilize MAE, MSE, or other perceptual losses to optimize a restoration
model. Such losses consider only the visual quality. However, the focuses of
human eyes and machines are different \cite{xie2017adversarial,tan2018feature}.
If an image is restored by optimizing only visual quality, the restored image
may not suitable for the task of tampering localization. Therefore, it needs to
make a restored image as close as possible to its plain version in several
aspects at the same time. To this end, we introduce three different losses to
optimize the restoration module, including a pixel-level loss, an image-level
loss, and a forensics-oriented loss. The used losses are described in detail as
follows.

\subsubsection{Pixel-level loss}
As did in traditional restoration methods, we use a pixel-level MAE loss to
measure the distance between every single pixel in a restored image and the
corresponding plain image. The pixel-level MAE loss $\mathcal{L}_{M\!A\!E}$ is
given as
\begin{equation}
    \mathcal{L}_{M\!A\!E}=\frac{1}{mn} \sum_{i=1}^{m}\sum_{j=1}^{n} |\mathbf{I}_{i,j}^{P}-\mathbf{I}_{i,j}^{R}|,
\end{equation}
where $\mathbf{I}^{P}$ denotes the plain tampered image, $\mathbf{I}^{R}$
denotes the restored image output from the restoration module, and $m$ and $n$
denote the height and width of an image, respectively.

\subsubsection{Image-level loss}
The $\mathcal{L}_{M\!A\!E}$ mentioned above focusses on the restoration of each
individual pixel, but does not consider the overall statistical distribution of
an image. Inspired by adversarial training in the field of image generation, we
use an adversarial training strategy to make the restored image as close as
possible to the plain tampered image at the image level. We treat the
restoration module as a generator and use a discriminator to classify the
restored image and the plain image. The discriminator in DCGAN
\cite{radford2015unsupervised} is adopted here. The generative and
discriminative losses are formulated as
\begin{equation}
    \mathcal{L}_{G} = -\log f(\mathbf{I}^{R};\theta_{D}),
\end{equation}
\begin{equation}
    \mathcal{L}_{D} = -\log f(\mathbf{I}^{P};\theta_{D})-\log(1-f(\mathbf{I}^{R};\theta_{D})),
\end{equation}
where $\theta_{D}$ denotes the parameters of the discriminator,
$f(\mathbf{I}^{R};\theta_{D})$ and $f(\mathbf{I}^{P};\theta_{D})$ represent the
probabilities output from the discriminator when the inputs are $\mathbf{I}^{R}$ and
$\mathbf{I}^{P}$, respectively.

\subsubsection{Forensics-oriented loss}
\label{Forensics-oriented loss}
Since the purpose of restoration in ReLoc is to make the tampering localization
module work better, we should ensure that the distributions of the tampered and
original pixels in the restored image are distinct enough, so that the tampering
localization module can distinguish them from each other after restoration. To
this end, the localization loss for training the localization module is also
used to optimize the restoration module. Specifically, the localization loss is
composed of the cross-entropy loss and the dice loss, which are defined as
\begin{equation}
\begin{split}
	\mathcal{L}_{C\!E}=-\frac{1}{mn} \sum_{i=1}^{m}\sum_{j=1}^{n}[\mathbf{G}_{i,j}\log(\mathbf{P}_{i,j})\\+(1-\mathbf{G}_{i,j})\log(1-\mathbf{P}_{i,j})],
\end{split}
\end{equation}
\begin{equation}
	\mathcal{L}_{D\!I\!C\!E}=1-\frac{2\sum_{i=1}^{m}\sum_{j=1}^{n}\mathbf{P}_{i,j}*\mathbf{G}_{i,j}}{\sum_{i=1}^{m}\sum_{j=1}^{n} \mathbf{P}_{i,j}^{2}+\sum_{i=1}^{m}\sum_{j=1}^{n}\mathbf{G}_{i,j}^{2}+\epsilon },
\end{equation}
where $\mathbf{G}$ is the ground-truth map, $\mathbf{P}$ is the predicted
probability map output by the localization module, and $\epsilon$ is a small
constant to avoid zero division. The total localization loss is a weighted sum of
$\mathcal{L}_{C\!E}$ and $\mathcal{L}_{D\!I\!C\!E}$:
\begin{equation}
	\mathcal{L}_{L}=\lambda_{1} \mathcal{L}_{C\!E} +(1-\lambda_{1}) \mathcal{L}_{D\!I\!C\!E},
\end{equation}
where $\lambda_{1}$ represents the weighting parameter.
\begin{figure}[tbp]
	\centering
	\includegraphics[width=0.98\columnwidth]{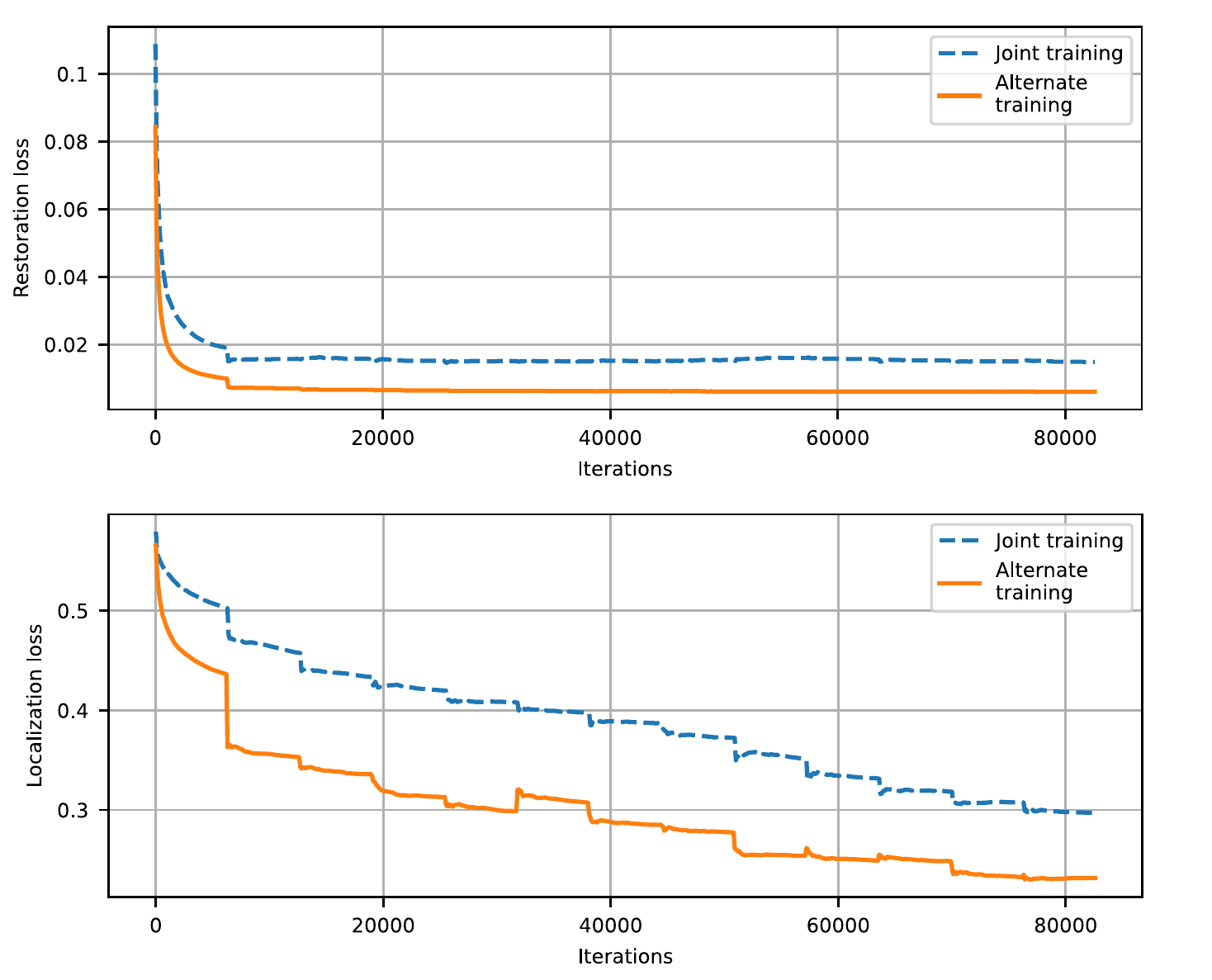}
	\caption{The training restoration and localization losses for joint training and alternate training.}
	\label{fig:TrainingLoss}
\end{figure}

On the whole, we incorporate $\mathcal{L}_{L}$ with $\mathcal{L}_{M\!A\!E}$ and
$\mathcal{L}_{G}$ to optimize the restoration module. The total restoration loss
is given by
\begin{equation}
\mathcal{L}_{R}=\lambda_{2} \mathcal{L}_{M\!A\!E}+\lambda_{3} \mathcal{L}_{G}+\lambda_{4} \mathcal{L}_{L},
\end{equation}
where $\lambda_{2}$, $\lambda_{3}$, and $\lambda_{4}$ are the weights for
$\mathcal{L}_{M\!A\!E}$, $\mathcal{L}_{G}$ and  $\mathcal{L}_{L}$, respectively.

\begin{algorithm}[t]
	\begin{small}
	\caption{The Optimization Algorithm}  
	\label{alg:opt_method}
	\begin{algorithmic}[1] 
		\Require \hspace{0.78ex}
		\begin{tabular}[t]{ll}
			$\mathbb{D}$: & training dataset; \\
			$t$:          & total training epochs; \\
			$\ell_{R}$:   & learning rate of restoration module; \\
			$\ell_{D}$:   & learning rate of discriminator; \\
			$\ell_{L}$:   & learning rate of localization module. \\
		\end{tabular} 
		\vspace{0.5em}
		\Ensure
		\begin{tabular}[t]{ll}
			$\theta_{R}$: & trained restoration module; \\
			$\theta_{L}$: & trained localization module. \\
		\end{tabular} 
		\vspace{0.5em}
		\State Randomly initialize $\theta_{R}$ and $\theta_{D}$, and initialize $\theta_{L}$ \newline with the weights in $\mathcal{M}^{P}$.
		\For {$epoch = 1,2,...,t$}
		\If{$epoch \bmod 2  \ne 0$}{
			\For {minibatch $(x_{i}^{P}, x_{i}^{D}, y_{i}) \subset \mathbb{D}$}
			\State $g_{\theta_{D}} \gets \nabla_{\theta_{D}}\mathcal{L}_{D} $
			\State $\theta_{D} \gets \theta_{D}-\ell_{D}\cdot g_{\theta_{D}}$ 
			
			\State $g_{\theta_{R}} \gets \nabla_{\theta_{R}}\mathcal{L}_{R} $
			\State $\theta_{R} \gets \theta_{R}-\ell_{R}\cdot g_{\theta_{R}}$ 
			\EndFor
		}
		\Else
		{
			\For {minibatch $(x_{i}^{P}, x_{i}^{D}, y_{i}) \subset \mathbb{D}$}
			\State $g_{\theta_{L}} \gets \nabla_{\theta_{L}}\mathcal{L}_{L} $
			\State $\theta_{L} \gets \theta_{L}-\ell_{L}\cdot g_{\theta_{L}}$ 
			\EndFor
		}
		\EndIf
		\EndFor
	\end{algorithmic}
	\end{small}
\end{algorithm} 

\subsection{Optimization Strategy}
\label{subsec:opt}


As the functions of the restoration module and the localization module in ReLoc
are different, an appropriate optimization strategy is important for training
the whole framework. Ideally, we can improve the training efficiency of the
whole framework by initializing $\mathcal{M}^{ReLoc}_{L}$ with the parameters of
$\mathcal{M}^{P}$ and only training $\mathcal{M}^{ReLoc}_{R}$. However, since
image restoration is ill-posed, we cannot perfectly restore the distorted
tampered images to plain images. Therefore, this training strategy would lead to
sub-optimal performance. To achieve better performance, it is better to optimize
both the restoration module and the localization module. Intuitively, one can
optimize the restoration and localization modules simultaneously in a joint
training manner. However, we found this strategy did not work well. The reason
is that the restoration module is in a relatively poor state in the early stage
of training, and the localization module would be misled by the poor restored
images produced in the early stage, finally resulting in poor localization
performance. To verify our analysis, we plotted the training restoration and
localization losses in Fig.~\ref{fig:TrainingLoss}. As shown in this figure,
when the two modules are jointly optimized, the localization loss decreases
slowly and takes more iterations to converge. This is due to the fact that the
restoration loss is relatively large in the early stage, and the poor restored
images make the localization module learn undesirable feature representations. 

To better optimize the two modules in ReLoc, we should make both of them in a
relatively good state. Therefore, we alternately optimize the restoration module
and the localization module, the optimization algorithm is shown in
Algorithm~\ref{alg:opt_method}. Specifically, we optimize either the restoration
module or the localization module in one epoch. At the first, we start to
optimize the restoration module. We firstly use the discriminator to classify
the plain images and the restored images, and optimize the discriminator based
on $\mathcal{L}_{D}$. Then, we optimize the restoration module with the guidance
of $\mathcal{L}_{R}$. After training the discriminator and the restoration
module with the whole dataset, we start another epoch and only optimize the
localization module with $\mathcal{L}_{L}$. The above process is repeated until
the whole model becomes convergent. As shown in Fig.~\ref{fig:TrainingLoss},
compared to joint training, the localization loss in alternate training
decreases faster and converges earlier. According to our experiments, by
optimizing the restoration and localization modules in such an alternate way,
the training process of ReLoc is more stable, and the model can achieve better
localization performance and robustness (please refer to
Table~\ref{tab:ablation_optimization}).

\begin{table}[t]
	\renewcommand{\arraystretch}{1.2}
	\caption{Localization performance of different restoration losses.}
	\label{tab:ablation_restore_loss}
	\centering
	\begin{tabular}{cccc}
		\toprule
		& F1 & IOU & AUC  \\ \midrule
		
		$\mathcal{L}_{M\!A\!E}$ & 0.515 &0.396  &0.940  \\
		$\mathcal{L}_{M\!A\!E}+\mathcal{L}_{G}$ & 0.523 & 0.404 & 0.944\\ 
		$\mathcal{L}_{M\!A\!E}+\mathcal{L}_{L}$ & 0.542 & 0.419  & \textbf{0.958}  \\
		
		$\mathcal{L}_{M\!A\!E}+\mathcal{L}_{G} +\mathcal{L}_{L}$& \textbf{0.567} & \textbf{0.444} & 0.955  \\ 
		\bottomrule
	\end{tabular}
\end{table}

\section{Experiments}
\label{Sec:experiments}
In this section, we evaluate the effectiveness of the proposed method. As JPEG
compression is one of the most commonly used post-processing operations, we
consider it as a typical example of post-processing and
investigate the robustness of tampering localization against JPEG compression.

\subsection{Experimental Setup}

\subsubsection{Datasets\label{subsubSec:dataset}} Three tampering datasets are
used for performance evaluations.
\begin{itemize}
\item \emph{Certificate PS dataset}. This dataset is generated from 4,840 original
certificate images that were captured by 77 different mobile phones. There are
five commonly used tampering operations for certificate images, \ie,
splicing, copy-move, removal, text addition, and text replacement. We invited 25
experts to choose one of the five operations to tamper with every original image.
The resulting tampered images are saved in uncompressed PNG format. 

\item \emph{DEFACTO dataset} \cite{mahfoudi2019defacto}. The tampered images in
this dataset are created from the MS-COCO dataset \cite{lin2014microsoft} in an
automatic way. Various common tampering operations are used to generate the
tampered images, including copy-move, splicing, and removal. In our experiments,
we randomly selected 98,000 tampered images from this dataset.

\item \emph{IMD2020 dataset} \cite{novozamsky2020imd2020}. This dataset consists
of 35,000 original images and 70,000 tampered images generated by GAN and
inpainting methods. In addition, it also contains 2,010 tampered images
collected from real scenes. We used the 2,010 realistic tampered images in our
experiments.
\end{itemize}
All the aforementioned tampered images are regarded as plain images in the
experiments, and we applied JPEG compression to them to obtained the distorted
images. For each dataset, we randomly selected 75\% images for training
and used the left 25\% for testing.

\subsubsection{Backbone networks}
We used SwinConv-Unet\footnote{\url{https://github.com/cszn/SCUNet}}
\cite{zhang2022practical} as our restoration module, which has achieved good
image restoration performance via combining swin transformer and convolutional
block. As for the localization module, we implemented it with three
state-of-the-art tampering localization methods, including
DFCN\footnote{\url{https://github.com/ZhuangPeiyu/Dense-FCN-for-tampering-localization}}
\cite{zhuang2021image},
MVSS-net\footnote{\url{https://github.com/dong03/MVSS-Net}}
\cite{chen2021image}, and
SCSE-Unet\footnote{\url{https://github.com/HighwayWu/Tianchi-FFT2}}
\cite{wu2022robust}, so as to verify that ReLoc can universally improve the
robustness against post-processing.

\subsubsection{Performance metrics}
\label{subsub:details}
As image tampering localization is a pixel-level binary classification task, we
use some commonly used metrics for binary classification to evaluate the
performance of the proposed framework, including F1-score, IOU (Intersection
over Union), and AUC (Area Under the ROC Curve). For computing the F1-score and
IOU, we applied thresholding to the predictions of all images with a fix
threshold of 0.5.

\begin{table}[t]
	\renewcommand{\arraystretch}{1.2}
    \caption{Localization performance of different optimization strategies.}
    \label{tab:ablation_optimization}
    \centering
    \begin{tabular}{ccccc}
    \toprule
       & F1 & IOU & AUC  \\ \midrule
    
       Joint training & 0.499 & 0.373 & 0.938 \\
    
    Alternate training  & \textbf{0.567} & \textbf{0.444}  &\textbf{0.955}   \\ 
    \bottomrule
    \end{tabular}
\end{table}

\subsubsection{Implementation details and evaluation protocol}
We implemented the proposed method with PyTorch 1.9.0 and ran all the
experiments with an NVIDIA V100 GPU\footnote{Code available at:
\url{https://github.com/ZhuangPeiyu/ReLoc}}. 
In the training phase, we used 128$\times$128 image blocks for training. Due to
the limitation of GPU memory, we set the batch size as large as possible for
different localization networks, namely, 56 for DFCN, 40 for SCSE-Unet, and 48
for MVSS-net. The restoration and localization modules were optimized with Adam
optimizer with the default setting in PyTorch. The initial learning rate was set
to $10^{-4}$ for both the restoration and localization modules. When the
validation loss did not descend in two consecutive epochs, we decreased the
learning rate by a factor of 0.8. For the Certificate PS dataset, $\lambda_{1}$,
$\lambda_{2}$, $\lambda_{3}$, and $\lambda_{4}$ were set to 0.2, 100, 1, and
0.05, respectively, while for DEFACTO and IMD2020 datasets,  $\lambda_{1}$,
$\lambda_{2}$, $\lambda_{3}$, and $\lambda_{4}$ were set to 0.2, 100, 1, and
0.1, respectively. 
During the testing phase, some high-resolution images could not be tested
directly due to memory limitation. Hence, we used the sliding window strategy in
all experiments. The window size was 512$\times$512 and the sliding step was
512. We combined the testing results of all image blocks within an image to
compute the performance metrics.

\begin{table*}[tbp]
	\renewcommand{\arraystretch}{1.75}
	\caption{Localization performance for three datasets in different training/testing situations. The first situation (gray background) shows the results for plain images and can be regarded as the performance upper bound of each method. The rest situations show the results for distorted images, where the best results are in bold.}
	\label{tab:single_QF_result}
	\centering
	\begin{tabular}{ccccccccccc}
		\toprule
		&  & \multicolumn{3}{c}{Certificate PS} & \multicolumn{3}{c}{DEFACTO} & \multicolumn{3}{c}{IMD2020} \\ \cline{3-11} 
		\multirow{-2}{*}{Localization methods} & \multirow{-2}{*}{Sistuations} & F1 & IOU & AUC & F1 & IOU & AUC & F1 & IOU & AUC \\ \midrule
		\rowcolor[gray]{0.95} \cellcolor{white}  \multirow{4}{*}{DFCN} 
		& $\mathcal{M}^{P}\{\mathbf{I}^P\}$ & 0.912 & 0.858 & 0.993 & 0.556 & 0.484 & 0.950 & 0.352 & 0.245 & 0.859 \\
		& $\mathcal{M}^{P}\{\mathbf{I}^D\}$ & 0.062 & 0.034 & 0.626 & 0.026 & 0.019 & 0.715 & 0.259 & 0.171 & 0.808 \\
		& $\mathcal{M}^{D|P}\{\mathbf{I}^D\}$ & 0.484 & 0.369 & 0.905 & 0.404 & 0.338 & 0.880 & 0.281 & 0.182 & 0.834 \\
		& $\mathcal{M}^{ReLoc}\{\mathbf{I}^D\}$ &  \textbf{0.567} & \textbf{0.444} & \textbf{0.955} & \textbf{0.429} & \textbf{0.359} & \textbf{0.888} & \textbf{0.329} & \textbf{0.223} & \textbf{0.849} \\ \midrule
		\rowcolor[gray]{0.95} \cellcolor{white}  \multirow{4}{*}{SCSE-Unet} 
		& $\mathcal{M}^{P}\{\mathbf{I}^P\}$ & 0.925 & 0.878 & 0.996 & 0.706 & 0.635 & 0.978 & 0.501 & 0.389 & 0.921 \\
		& $\mathcal{M}^{P}\{\mathbf{I}^D\}$ & 0.031 & 0.019 & 0.587 & 0.281 & 0.238 & 0.896 & 0.391 & 0.296 & 0.879 \\ 
		& $\mathcal{M}^{D|P}\{\mathbf{I}^D\}$ & 0.604 & 0.484 & 0.949 & 0.582 & 0.516 & 0.949 & 0.416 & 0.306 & 0.898 \\
		& $\mathcal{M}^{ReLoc}\{\mathbf{I}^D\}$ & \textbf{0.651} & \textbf{0.538} & \textbf{0.962} & \textbf{0.611} & \textbf{0.543} & \textbf{0.953} & \textbf{0.454} & \textbf{0.340} & \textbf{0.915} \\ \midrule
		\rowcolor[gray]{0.95} \cellcolor{white}  \multirow{4}{*}{MVSS-net} 
		& $\mathcal{M}^{P}\{\mathbf{I}^P\}$ & 0.789 & 0.687 & 0.993 & 0.558 & 0.478 & 0.943 & 0.384 & 0.280 & 0.838 \\
		& $\mathcal{M}^{P}\{\mathbf{I}^D\}$ & 0.011 & 0.006 & 0.642 & 0.161 & 0.126 & 0.849 & 0.228 & 0.161 & 0.756 \\
		& $\mathcal{M}^{D|P}\{\mathbf{I}^D\}$ & 0.391 & 0.270 & 0.916 & 0.507 & 0.429 & 0.910 & 0.314 & 0.216 & \textbf{0.819} \\
		& $\mathcal{M}^{ReLoc}\{\mathbf{I}^D\}$ & \textbf{0.410} & \textbf{0.288} & \textbf{0.942} & \textbf{0.515} & \textbf{0.438} & \textbf{0.917} & \textbf{0.323} & \textbf{0.228} & 0.810 \\
		\bottomrule
	\end{tabular}
\end{table*}

\subsection{Ablation Study}
In this subsection, we evaluate the effectiveness of the design of ReLoc through
ablation experiments. Two key factors that would affect the performance of ReLoc
has been investigated, \ie, the restoration loss and optimization strategy for
training the two modules. Note that the experiments were conducted on the
Certificate PS dataset by using DFCN as the localization module.

\subsubsection {Restoration loss}
To investigate the impact of different restoration losses on the performance of
ReLoc, we used different losses and their combinations to optimize the
restoration module. The experimental results are shown in
Table~\ref{tab:ablation_restore_loss}. We can observe that compared with using
only $\mathcal{L}_{M\!A\!E}$, adding $\mathcal{L}_{G}$ is effective to help the
restoration module to produce better restored images. The F1-score is improved
from 0.515 to 0.523. On the other hand, introducing $\mathcal{L}_{L}$ to
optimize the restoration module is also beneficial for improving the
performance. The F1-score is increased from 0.515 to 0.542 in this case. This
implies that by introducing the localization loss, the restoration module can be
guided to pay more attention to strengthening the differences between the
original pixels and the tampered pixels. In such a way, the distorted tampering
traces can be re-enhanced. Finally, by combining $\mathcal{L}_{M\!A\!E}$,
$\mathcal{L}_{G}$, and $\mathcal{L}_{L}$ together, the restoration module can
work better, and the F1-score is improved from 0.542 to 0.567.

\subsubsection {Optimization strategy}
To evaluate which strategy is better for optimizing the two modules in ReLoc, we
separately conducted experiments to train the two modules jointly and
alternately. It can be seen from Table~\ref{tab:ablation_optimization} that by
alternately training the restoration and localization modules, the obtained
performance is better than training them jointly. The net increases of F1-score,
IOU, and AUC are 0.068, 0.071, and 0.017, respectively.

\subsection {Robustness against Fixed JPEG Compression}

For a given JPEG image, its compression quality factor (or quantization table)
is available in the JPEG file. To perform tampering localization, ideally we can
build a matched detection model by collecting training images with the same
quality factor. Therefore, in this subsection we evaluate the robustness of the
proposed framework against a fixed JPEG compression, where the JPEG quality
factor (QF) is set to 75. 
We separately used DFCN, SCSE-Unet, and MVSS-net as the localization module in ReLoc, and
conducted experiments in four different training/testing situations. The result
are shown in Table~\ref{tab:single_QF_result}. From this table, we mainly
obtain two observations.

\begin{table*}[tbp]
	\renewcommand{\arraystretch}{1.75}
	\caption{The localization performance regarding to different JPEG compressions.}
	\label{tab:multiQF_performance}
	\centering
	\begin{tabular}{cccccccccccc}
		\toprule
		\multirow{2}{*}{Dataset} & \multirow{2}{*}{Localization methods} & \multirow{2}{*}{Sistuations} &  & QF60 &  &  & QF70 &  &  & QF80 &  \\ \cline{4-12} 
		&  &  & F1 & IOU & AUC & F1 & IOU & AUC & F1 & IOU & AUC \\ \midrule
		\multirow{6}{*}{Certificate PS} & \multirow{2}{*}{DFCN} & $\mathcal{M}^{D|P}\{\mathbf{I}^D\}$ & 0.443 & 0.335 & 0.880 & 0.466 & 0.357 & 0.891 & 0.494 & 0.381 & 0.903 \\
		&  & $\mathcal{M}^{ReLoc}\{\mathbf{I}^D\}$ &  \textbf{0.479} & \textbf{0.370} &\textbf{0.900}  &\textbf{0.513}  & \textbf{0.396} &\textbf{0.938}  &\textbf{0.584}  & \textbf{0.470} & \textbf{0.957}  \\ \cline{2-12} 
		& \multirow{2}{*}{SCSE-Unet} & $\mathcal{M}^{D|P}\{\mathbf{I}^D\}$ & 0.509 & 0.391 & 0.928 & 0.531 & 0.407 & 0.939 & 0.578 & 0.453 & 0.951 \\
		&  & $\mathcal{M}^{ReLoc}\{\mathbf{I}^D\}$ & \textbf{0.510} &\textbf{0.391}  & \textbf{0.930} & \textbf{0.609} &\textbf{0.486}  &\textbf{0.955}  &\textbf{0.700} &\textbf{0.589}  &\textbf{0.969}  \\ \cline{2-12} 
		& \multirow{2}{*}{MVSS-net} & $\mathcal{M}^{D|P}\{\mathbf{I}^D\}$ & 0.328  & 0.216 & 0.874 & 0.337 & 0.222 & 0.886 & 0.348 & 0.231 & 0.899 \\
		&  & $\mathcal{M}^{ReLoc}\{\mathbf{I}^D\}$ &\textbf{0.337} & \textbf{0.222} & \textbf{0.893} & \textbf{0.374} & \textbf{0.254} & \textbf{0.920} & \textbf{0.441} & \textbf{0.316} & \textbf{0.938} \\ \midrule
		\multirow{6}{*}{DEFACTO} & \multirow{2}{*}{DFCN} & $\mathcal{M}^{D|P}\{\mathbf{I}^D\}$ & 0.402 & 0.338 & 0.889 & 0.409 & 0.345 & 0.890 & 0.463 & 0.398 & 0.907 \\
		&  & $\mathcal{M}^{ReLoc}\{\mathbf{I}^D\}$ &\textbf{0.421} & \textbf{0.354} & \textbf{0.896} & \textbf{0.432} &\textbf{0.365}  & \textbf{0.896} & \textbf{0.485} & \textbf{0.415} & \textbf{0.910} \\ \cline{2-12}  
		& \multirow{2}{*}{SCSE-Unet} & $\mathcal{M}^{D|P}\{\mathbf{I}^D\}$ & 0.588 & 0.518 & \textbf{0.951} & 0.594 & 0.525 & \textbf{0.953} & 0.629 & 0.562 & \textbf{0.959} \\
		&  & $\mathcal{M}^{ReLoc}\{\mathbf{I}^D\}$ & \textbf{0.600} & \textbf{0.533} & 0.950 & \textbf{0.604} & \textbf{0.538} & 0.952 & \textbf{0.634} & \textbf{0.567} & 0.958 \\ \cline{2-12}  
		& \multirow{2}{*}{MVSS-net} & $\mathcal{M}^{D|P}\{\mathbf{I}^D\}$  & 0.461 & 0.392 & \textbf{0.914} & 0.467 & 0.398 & \textbf{0.916} & 0.507 & 0.432 & 0.900 \\
		&  & $\mathcal{M}^{ReLoc}\{\mathbf{I}^D\}$ & \textbf{0.477} & \textbf{0.404} & 0.910 & \textbf{0.489} & \textbf{0.415} & 0.911 & \textbf{0.532} & \textbf{0.455} & \textbf{0.925} \\ \midrule
		\multirow{6}{*}{IMD2020} & \multirow{2}{*}{DFCN} & $\mathcal{M}^{D|P}\{\mathbf{I}^D\}$ & 0.287 & 0.186 & 0.829 & 0.289 & 0.188 & 0.835 & 0.290 & 0.189 & 0.836 \\
		&  & $\mathcal{M}^{ReLoc}\{\mathbf{I}^D\}$ & \textbf{0.338} & \textbf{0.229} & \textbf{0.841} & \textbf{0.344} &\textbf{0.233}  &\textbf{0.854}  &\textbf{0.357} &\textbf{0.244}  & \textbf{0.861} \\ \cline{2-12}
		& \multirow{2}{*}{SCSE-Unet} & $\mathcal{M}^{D|P}\{\mathbf{I}^D\}$ & 0.429 & 0.320 & \textbf{0.897} & 0.429 & 0.321 & 0.900 & 0.436 & \textbf{0.356} & 0.903 \\
		&  & $\mathcal{M}^{ReLoc}\{\mathbf{I}^D\}$ &\textbf{0.446}  &\textbf{0.334}  & \textbf{0.897} & \textbf{0.454} & \textbf{0.340} & \textbf{0.901} & \textbf{0.459} & 0.347 &\textbf{0.905}  \\ \cline{2-12}
		& \multirow{2}{*}{MVSS-net} & $\mathcal{M}^{D|P}\{\mathbf{I}^D\}$ & 0.320 & 0.224 & \textbf{0.813} & 0.331 & \textbf{0.234} & \textbf{0.814} & 0.323 & \textbf{0.231} & \textbf{0.813} \\
		&  & $\mathcal{M}^{ReLoc}\{\mathbf{I}^D\}$ & \textbf{0.331} & \textbf{0.230} & 0.812 & \textbf{0.334} & 0.233 & \textbf{0.814} & \textbf{0.325} & 0.228 & 0.812\\ \bottomrule
	\end{tabular}
\end{table*}

Firstly, when a model trained with plain images is used to test distorted images
(\ie, the case $\mathcal{M}^{P}\{\mathbf{I}^D\}$), the localization performance
would be significantly degraded. This is not surprising as the tampering traces
are weak in distorted images. The performance decline for the Certificate PS
dataset is the worst, where the F1-scores are decreased close to 0. The reason
is that the plain tampered images in this dataset are uncompressed, and there
are considerable differences between the tampered and original regions. The
localization model can easily learn discriminative features from the plain
images, but the learned features are not suitable for resisting JPEG
compression. In the DEFACTO and IMD2020 datasets, some plain tampered images
have been undergone JPEG compression after tampering, so the models
$\mathcal{M}^{P}$ trained with these datasets could somewhat adapt to JPEG
compression and thus perform relatively better.

More importantly, we observe that using distorted tampered images to directly
train the localization network can improve the robustness against JPEG
compression to a certain extent, but the improvement is not as significant as
the use of ReLoc. For example, by using DFCN as the localization module, the
average improvement of F1-score between $\mathcal{M}^{D|P}\{\mathbf{I}^D\}$ and
$\mathcal{M}^{P}\{\mathbf{I}^D\}$ is 0.274, while the average improvement is
0.326 when using ReLoc. Similar results can be obtain by considering SCSE-Unet
and MVSS-net. On average, the improvements of F1-score between the cases
$\mathcal{M}^{ReLoc}\{\mathbf{I}^D\}$ and $\mathcal{M}^{D|P}\{\mathbf{I}^D\}$
are 0.052, 0.038, and 0.012, respectively for DFCN, SCSE-Unet, and MVSS-net.
Based on these experimental results, it can be concluded that the proposed
framework can indeed effectively improve the robustness of tampering
localization against fixed JPEG compression.

\subsection{Robustness against Multiple JPEG Compressions}
As we all know, in real-world scenarios, different tampered images are usually
subjected to different JPEG compressions. Training a specific ReLoc model
for each JPEG compression under investigation is impractical because it would be
time-consuming and the implementations of JPEG compression would vary from
manufacturers and software. Therefore, in this subsection, we evaluate the
robustness against different JPEG compressions by using a single model.

To generate distorted images for model training, we sampled uniformly the JPEG
QFs between 70 and 100, similar to what was done in \cite{wu2022robust}. In the
testing phase, three different QFs were considered, \ie, 60, 70, and 80. Two
training/testing situations, $\mathcal{M}^{D|P}\{\mathbf{I}^D\}$ and
$\mathcal{M}^{ReLoc}\{\mathbf{I}^D\}$, were involved in this experiment. The
experimental results are shown in Table~\ref{tab:multiQF_performance}. From this
table, we observe that no matter which localization method was adopted, the
localization performance of $\mathcal{M}^{ReLoc}\{\mathbf{I}^D\}$ is better than
that of $\mathcal{M}^{D|P}\{\mathbf{I}^D\}$. When the testing QF is 80, the
improvements of F1-score averaged over three datasets are 0.060, 0.050, and
0.040 for DFCN, SCSE-Unet, and MVSS-net, respectively. The average improvements
for QF 70 are relatively slighter, which are 0.042, 0.037, and 0.021 for DFCN,
SCSE-Unet, and MVSS-net, respectively. When the testing QF is unseen in the
training phase (\ie, QF 60), the average improvements achieved by ReLoc for
DFCN, SCSE-Unet, and MVSS-net are 0.035, 0.010, and 0.012, respectively, which
are still considerable. These experimental results indicate that ReLoc is also
effective for improving the robustness against multiple JPEG compressions.

\begin{figure*}[htbp]
	\centering
	\includegraphics[width=\textwidth]{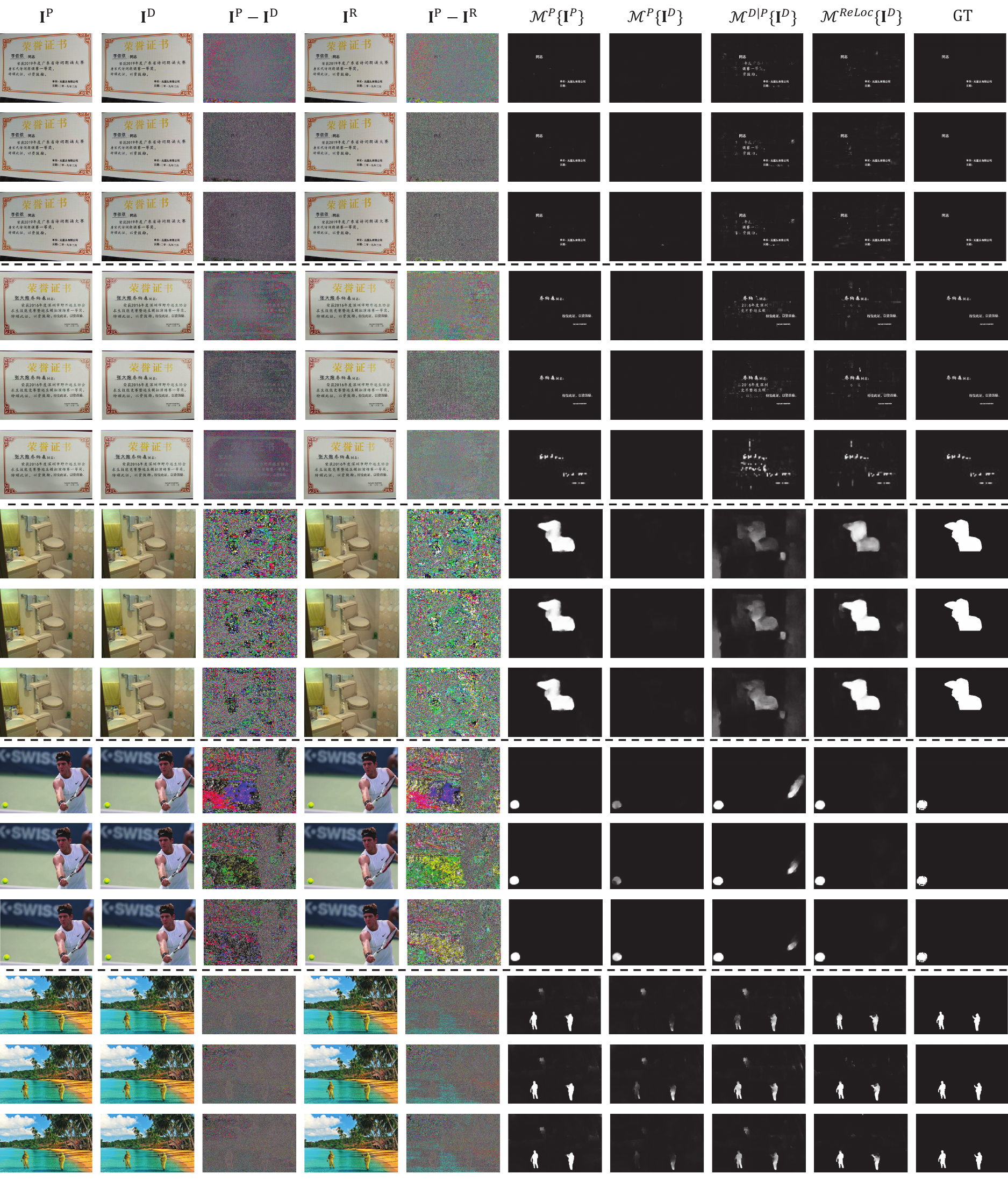}
	\caption{Examples of tampering localization results in different situations. Each
	super-row (separated by dashed lines) corresponds to an example. In each
	example, the distorted images $\mathbf{I}^D$ from top to bottom were compressed with JPEG
	quality factors 60, 70, and 80, respectively. Examples \#1 and \#2 are from
	the Certificate PS dataset, examples \#3 and \#4 are from the DEFACTO
	dataset, while the last example is from the IMD2020 dataset.}
	\label{fig:LocalizationResults}
\end{figure*}

\subsection{Qualitative Comparisons}
In order to assess the performance intuitively, we show some examples of
tampering localization results in Fig.~\ref{fig:LocalizationResults}. In each
example in this figure, from top to bottom, $\mathbf{I}^D$ was generated through
compressing $\mathbf{I}^P$ with QFs 60, 70, and 80, respectively. By comparing
the results of $\mathcal{M}^{P}\{\mathbf{I}^P\}$ and
$\mathcal{M}^{P}\{\mathbf{I}^D\}$, we can see that the latter case would lead to
much more false alarms and/or missed detections. Although this phenomenon can be
mitigated via fine-tuning the localization network with distorted images
(\ie, $\mathcal{M}^{D|P}\{\mathbf{I}^D\}$), there are still more false predictions
compared to the proposed method (\ie, $\mathcal{M}^{ReLoc}\{\mathbf{I}^D\}$). For
instance, the tampered region in the fourth example is the tennis ball. The
localization result is correct in the $\mathcal{M}^{P}\{\mathbf{I}^P\}$
situation. However, there are missed detections in the
$\mathcal{M}^{P}\{\mathbf{I}^D\}$ situation, especially when the QF is low. In
the $\mathcal{M}^{D|P}\{\mathbf{I}^D\}$ situation, missed detections are
decreased, but false alarms are introduced. By contrast, in the
$\mathcal{M}^{ReLoc}\{\mathbf{I}^D\}$ situation, via employing the ReLoc
framework, we can accurately locate the tampered region for different QFs.

\subsection{Transferability of the Restoration Module} 
In this subsection, we evaluate the transferability of restoration module of a
well-trained ReLoc model. We firstly trained the restoration module together
with a localization module using the proposed method. Then, we deployed the
trained restoration module with another localization model trained with plain
tampered images ($\mathcal{M}^{P}$). We aim to evaluate whether the performance
of another localization model would be improved or not by feeding it the images output
by the restoration module.

As shown in Table~\ref{tab:changeRestoration}, compared to the results in the
case $\mathcal{M}^{P}\{\mathbf{I}^D\}$ (see Table~\ref{tab:single_QF_result}),
when directly combining the restoration module trained along with DFCN and the
SCSE-Unet localization model, the F1-score of SCSE-Unet is improved from 0.031
to 0.441 on the Certificate PS dataset and improved from 0.281 to 0.612 on the
DEFACTO dataset. The same phenomenon can be observed when the restoration module
trained along with SCSE-Unet was deployed with the DFCN. the improvements of
F1-score are 0.342 and 0.267 on the Certificate PS dataset and the DEFACTO
dataset, respectively. Such experimental results have verify that the
restoration module in ReLoc is transferable.

\begin{table}[tbp]
	\renewcommand{\arraystretch}{1.5}
	\caption{The localization results obtained by ReLoc via replacing the
	localization module ($\mathcal{M}^{ReLoc}_L$) in a well-trained ReLoc model with
	another localization module trained with plain images ($\mathcal{M}^{P}$).
	The columns $\mathcal{M}^{ReLoc}_L$ and $\mathcal{M}^{P}$ denote the types of network
	structures. The values in parentheses are the
	improvements compared to the case $\mathcal{M}^{P}\{\mathbf{I}^D\}$.}
	\centering
	\label{tab:changeRestoration}
	\scalebox{0.9}{
	\begin{tabular}{C{1.5cm}C{1.32cm}C{1.32cm}ccc}
		\toprule
		Dataset & $\mathcal{M}^{ReLoc}_L$ & $\mathcal{M}^{P}$ & F1 & IOU & AUC \\
		\midrule
		\multirow{3}{*}{Certificate PS} 
		& DFCN & SCSE-Unet & \tabincell{c}{0.441 \\ (+0.410)} &	\tabincell{c}{0.349 \\ (+0.330)} & \tabincell{c}{0.844 \\ (+0.257)} \\ \cline{2-6} 
		& SCSE-Unet & DFCN & \tabincell{c}{0.404 \\ (+0.342)} & \tabincell{c}{0.309 \\ (+0.275)} & \tabincell{c}{0.891 \\ (+0.265)} \\
		\midrule
		\multirow{3}{*}{DEFACTO} 
		& DFCN & SCSE-Unet & \tabincell{c}{0.612 \\ (+0.331)} & \tabincell{c}{0.543 \\ (+0.305)} & \tabincell{c}{0.953 \\ (+0.057)} \\ \cline{2-6}
		& SCSE-Unet & DFCN & \tabincell{c}{0.293 \\ (+0.267)} & \tabincell{c}{0.248 \\ (+0.229)} & \tabincell{c}{0.859 \\ (+0.144)} \\
		\bottomrule
	\end{tabular}}
\end{table}

\section{Conclusion}
\label{Sec:conclusion}
In this paper, in order to improve the robustness of tampering localization
against post-processing, we propose a restoration-assisted framework named
ReLoc. The ReLoc framework is composed of an image restoration module and a
tampering localization module. We utilize the image restoration module to
recover tampering traces from distorted images. By optimizing the restoration
module with pixel-level, image-level, and forensics-oriented losses, it is able
to produce a restored image that is closer to the plain image, which can help
the localization module learn more discriminative and robust features. By
adopting an alternate training strategy, we make the training process more
stable and further improve the localization performance. Via considering JPEG
compression as a typical example of post-processing, we have conducted extensive
experiments to evaluate the effectiveness of ReLoc. The experimental results
show that ReLoc can significantly improve the robustness of tampering
localization. Moreover, our experiments show that the trained restoration module
is transferable, meaning that it can be individually and flexibly deployed with
other tampered localization methods.

In the future, we will further study to employ ReLoc for improving the
robustness against other post-processing operations. For example, blurring,
scaling, noise corruption, \etc. On the other hand, the restoration module and
the localization module in ReLoc are flexible, we will update the modules with
newly proposed and more effective restoration/localization methods to further
improve the robustness against various post-processing operations.

\bibliographystyle{IEEEtran}
\bibliography{refs}

\end{document}